\documentclass[twoside]{article}

\usepackage{PRIMEarxiv}

\usepackage{booktabs} % For formal tables
\usepackage{graphicx} %NUESTRO
\usepackage{latexsym} %NUESTRO
\usepackage{multirow} % NUESTRO
\usepackage{url} 

\usepackage{amsmath,amssymb,amsfonts} %NUESTRO
\usepackage{algorithm,algpseudocode} %NUESTRO
\usepackage{xcolor} %NUESTRO
\usepackage{caption}
\usepackage{subcaption}

\begin{document}
\title{Analysis of voice recordings features for
Classification of Parkinson's Disease}

\author{Beatriz Pérez-Sánchez, Noelia Sánchez-Maroño, Miguel A. Díaz-Freire \\
    Universidade da Coruña, CITIC \\
    Campus de Elviña, s/n \\
    15071, A Coruña, Spain  \\
    \texttt{\{beatriz.perezs, noelia.sanchez, m.diaz.freire\}@udc.es}
}

\pagestyle{fancy}
\thispagestyle{empty}
\rhead{ \textit{ }}

\fancyhead[LO]{Analysis of voice recordings features for Classification of Parkinson's Disease}
\fancyhead[RE]{Beatriz Pérez-Sánchez et al.}

\maketitle

\begin{abstract}
Parkinson's disease (PD) is a chronic neurodegenerative disease. Early diagnosis is essential to mitigate the progressive deterioration of patients' quality of life. The most characteristic motor symptoms are very mild in the early stages, making diagnosis difficult. Recent studies have shown that the use of patient voice recordings can aid in early diagnosis. Although the analysis of such recordings is costly from a clinical point of view, advances in machine learning techniques are making the processing of this type of data increasingly accurate and efficient. Vocal recordings contain many features, but it is not known whether all of them are relevant for diagnosing the disease.

This paper proposes the use of different types of machine learning models combined with feature selection methods to detect the disease. The selection techniques allow to reduce the number of features used by the classifiers by determining which ones provide the most information about the problem. The results show that machine learning methods, in particular neural networks, are suitable for PD classification and that the number of features can be significantly reduced without affecting the performance of the models. 
\end{abstract}


% *****************************************************
\section{Introduction} \label{sec:Introduction}
% *****************************************************

Parkinson's disease (PD) is a chronic neurodegenerative disease that affects the central nervous system and is suffered by approximately 10 million people worldwide, being the second most common neurodegenerative disease after Alzheimer's \cite{hirsch2016incidence}. PD is characterized by the progressive degeneration of neurons in a specific region of the brain. This region, known as the \textit{substantia nigra}, is responsible, among other things, for producing one of the neurotransmitters that controls body movement: dopamine \cite{latif2021dopamine}. Between 70-80\% of the neurons responsible for the production of dopamine gradually deteriorate \cite{el2006detection}, which produces the most characteristic motor symptoms of the disease \cite{emamzadeh2018parkinson}, among which are tremor of the hands, arms, legs, jaw and face, stiffness of the extremities, slowness of movement and problems with balance and coordination \cite{hoehn2001parkinsonism}. However, PD also affects the production of other types of neurotransmitters such as serotonin, acetylcholine and norepinephrine  \cite{scatton1983reduction}. The deterioration in the neurons that produce these other types of neurotransmitters causes other symptoms such as speech alteration, loss of smell or sleep disorder \cite{pfeiffer2016non}.

 %\cite{poewe2017parkinson} referencia {b1} quitada por ir ahorrando 
 %\cite{b6} idem
 %\cite{b8} idem
 %\cite{b10} idem

Currently, the diagnosis of the disease is made through clinical evaluation of the characteristic motor symptoms \cite{jankovic2008parkinson}. However, being a progressive disease, the symptoms are mild in the initial stages and worsen over time. Furthermore, the presence of some symptoms or others is variable in each patient \cite{wolters2008variability} and other diseases present the same symptoms, making early diagnosis of PD difficult.
There are other diagnostic methods that are performed when the presence of the disease is suspected, such as the study of the dopamine transporter (\textit{DATScan}) \cite{marshall2003role}, Positron Emission Tomography or Magnetic Resonance Imaging (MRI) \cite{brooks2010imaging}, although they are less widespread due to their lack of effectiveness, apart from being costly in both time and resources and even invasive.

Recent studies have shown that symptoms related to speech disorders can be detected in the early stages of PD and, in addition, are present in approximately 90\% of patients \cite{aich2019supervised}. The analysis of voice recordings has proven to be relevant in the diagnosis of PD and turns out to be a non-invasive method with a much lower cost than those mentioned above. Because the clinical analysis of this type of records is difficult to understand, automatic methodologies are required for their evaluation. Machine learning techniques have demonstrated their validity to process data from voice recordings \cite{sakar2013collection}, \cite{ma2014efficient}.

Currently there is no standard methodology for the study of voice recordings and it is specifically unknown which characteristics or metrics are really decisive for the diagnosis of the disease. Thus, datasets of voice recordings from PD patients contain a large number of non-specific features that make the final diagnosis difficult. Feature selection (FS) methods allow us to reduce the number of features that classification models will process without drastically affecting their performance, so that we can get a more specific view of which features are actually necessary to determine the diagnosis \cite{kursun2012selection}.
The objective of this work is to find, thanks to FS, those characteristics or groups of characteristics present in the datasets of voice recordings of patients with PD that are most decisive for the classification between individuals affected by the disease and individuals healthy, so that the classification models are more generalizable and guide clinical professionals.

The work is structured as follows: Section \ref{sec:voice-recordings} describes characteristics present in the voice recordings. In  Section \ref{sec:state-of-art} the state of the art in the classification of PD and more specifically in the classification of PD using voice recordings will be studied. In Section \ref{sec:methodology} the work methodology will be defined, explaining from the dataset used to the evaluation metrics. Section \ref{sec:experimental-results} shows the results obtained and their discussion. Finally, in Section \ref{sec:conclusions} the conclusions of the study carried out will be presented.

% ************************************************************************
\section{Characteristics of voice recordings} \label{sec:voice-recordings}
% ************************************************************************
The diagnosis of PD through patients' speech alterations is made by analyzing their voice recordings which consist of recordings of patients performing different voice exercises such as speaking, pronouncing a vowel for a long time, etc. Audio samples are composed of different sound waves whose direct analysis is complex and from which obtaining relevant information becomes complicated. However, different metrics can be obtained from these recordings that characterize the patient's voice and provide relevant information when diagnosing different conditions. In general, the characteristics used focus on quantifying the variations in movement, vibration and noise compared to healthy individuals in the regions of the vocal cords, lips and mouth, as they are the main areas affected by PD. Taking into account the features present in the dataset used\footnote{\url{https://doi.org/10.24432/C5MS4X}. Retrieved on September 2023.} \cite{sakar2019comparative} (see section \ref{subsec:datasets}), we will differentiate the following groups:

\begin{enumerate}
    \item[\textbf{(G1)}] \textbf{Reference features}: focus on vocal fold oscillation patterns, such as \textit{jitter} and \textit{shimmer} variants, which measure irregularities in vocal fold vibration. 
    \item [\textbf{(G2)}] \textbf{Temporal Frequency Features}: These are extracted from spectrograms of speech signals and contain speech intensity, formant frequencies (intensity peaks in the sound spectrum), and bandwidth-based features.
    \item [\textbf{(G3)}] \textbf{Mel Frequency Cepstral Coefficients (MFCC)}: They emulate the filtering properties of the human ear. These coefficients can detect subtle changes in the movement of the tongue and lips\cite{drissi2019diagnosis}.
    \item [\textbf{(G4)}] \textbf{Wavelet Transform (WT)}: They quantify deviations in the fundamental frequency. This is particularly relevant, since sustained vowels show significant differences in periodicity between healthy and sick individuals \cite{oung2018empirical}.
    \item[\textbf{(G5)}] \textbf{Vocal folds features}:  these characteristics evaluate the periodicity of glottal closure through the glottal quotient, turbulent noise caused by incomplete closure of the vocal folds (glottal noise excitation), noise generated by pathological vibration of the vocal folds (vocal fold excitation ratio) and the noise derived from the decomposition of the speech signal into elementary components (empirical modal decomposition), among other measurements.
    \item [\textbf{(G6)}]\textbf{Tunable Q-factor wavelet transform (TQWT)}: the Q factor is an indicator of the selectivity of a filter. These features, taking advantage of the periodic properties discussed in WT-based features, adjust the parameters of the signals to achieve higher frequency resolution, which improves the ability of classifiers to distinguish between healthy and diseased individuals \cite{drissi2019diagnosis}. 
\end{enumerate}

% *****************************************************
\section{State of the art} \label{sec:state-of-art}
% *****************************************************
Machine learning methods have been widely used in the clinical area and have recently been successfully applied to the diagnosis of PD \cite{mei2021machine}. These diagnosis can be determined with very varied types of data. Some of these studies focus on the analysis of neuroimages \cite{choi2017refining}, while others use different data modalities such as electroencephalographic (EEG) waves \cite{oh2020deep}, handwritten dynamics \cite{pereira2018handwritten}, the indicators extracted from the patient's gait \cite{abdulhay2018gait} and the voice recordings \cite{wroge2018parkinson}.
diagnostic

Regarding the studies carried out on voice recordings, which are the ones that interest us for this work, We can find several datasets publicly accessible through the repository of the University of California Irvine (UCI). These sets differ from each other in the number of samples and features that compose them. We will focus one of the most recent sets, proposed by C. Sakar et al. \cite{misc_parkinson's_disease_classification_470} and consisting of 756 samples and 754 features. This dataset has been used in the following papers and obtained results are shown in Table \ref{tab:state-of-art-results}:

 \begin{itemize}
    \item \textit{Sakar et al.} \cite{sakar2019comparative} obtained the best results with minimum Redundancy Maximum Relevance (mRMR) as FS algorithm and Support Vector Machines (SVM) as classifier.
    \item \textit{Hoq et al.} \cite{hoq2021vocal} used SVM together with Principal Component Analysis (PCA) and an autoencoder.
    \item \textit{Demir et al.} \cite{demir2021feature} proposed a new method in which they transform voice recordings into images and then use a recurrent neural network (Long Short-Term Memory Network).
    \item \textit{Xiong and Lu} \cite{xiong2020deep} used the \textit{Grey Wolf Optimizer (GWO)} algorithm as a FS step and then an extraction was done using autoencoders. For classification they use up to six different models, among which are SVMs.
    \item \textit{Gunduz} \cite{gunduz2019deep} used convolutional neural networks using two different frameworks when passing the characteristics of the dataset to the network.
    \item \textit{Ashour et al.} \cite{ashour2020novel} used to reduce dimension PCA and selection by calculating eigenvector centrality. Subsequently, they used SVM as a classifier.
    \item \textit{Gunduz} \cite{gunduz2021efficient} used two filter-based FS algorithms such as ReliefF and Fisher Score, combined with autoencoders to finally classify using SVM multi-kernel.
    \item \textit{Solana-Lavalle and Rosas-Romero} \cite{solana2021analysis} applied PCA and FS based on wrappers as a previous step. Classification is then performed using kNN, multilayer perceptron (MLP), SVM and Random Forest (RF).
    \item \textit{Lambda et al.} \cite{lamba2022hybrid} used a hybrid approach for FS combining the mutual information gain algorithm and recursive feature elimination, to subsequently classify using RF and \textit{XGBoost}.
    \item \textit{Masud et al.} \cite{masud2021crowd} used a neural network with autoencoder together with the Crow Search Algorithm (CSA).
    \item \textit{Yuçelbas} \cite{yucelbacs2021new} reduced the number of features by Independent Component Analysis (ICA) and variants. RF is used in the classification phase.\\
\end{itemize}

\begin{table}[]\centering
\caption{Results obtained in other studies with the dataset used. NA: Not available. MCC: Matthew's Correlation Coefficient.}
\label{tab:state-of-art-results}
\begin{tabular}{|l|c|c|}
\hline
\multicolumn{1}{|c|}{\textbf{Research work}}  & \textbf{Accuracy} & \textbf{MCC} \\\hline
\textit{Sakar et al., 2019} \cite{sakar2019comparative} & 0.860 & 0.590 \\\hline
\textit{Hoq et al., 2021} \cite{hoq2021vocal} & 0.935 & 0.788 \\\hline
\textit{Demir et al., 2021} \cite{demir2021feature} & 0.943 & NA \\\hline
\textit{Xiong et al., 2020} \cite{xiong2020deep} & 0.95 & NA \\ \hline
\textit{Gunduz, 2019} \cite{gunduz2019deep} & 0.869 & 0.632 \\\hline
\textit{Ashour et al., 2020} \cite{ashour2020novel} & 0.94 & NA \\\hline
\textit{Gunduz, 2021} \cite{gunduz2021efficient} & 0.912 & 0.772 \\ \hline
\textit{Solana-Lavalle et al., 2021} \cite{solana2021analysis} & 0.947 & 0.868 \\\hline
\textit{Lambda et al., 2022} \cite{lamba2022hybrid} & 0.947 & NA \\ \hline
\textit{Masud et al., 2021} \cite{masud2021crowd} & 0.96 & NA \\ \hline
\textit{Yuçelbas, 2021} \cite{yucelbacs2021new} & 0.821 & NA \\ \hline
\end{tabular}\\
\end{table}

Regarding the models used to classify between individuals with PD and healthy individuals, we can say that SVMs are the most used, while ANNs have been gaining prominence and have shown to obtain good results. On the other hand, dimensionality reduction is a common preliminary step in all works.

% *****************************************************
\section{Methodology} \label{sec:methodology}
% *****************************************************
This section describes the dataset used (Section \ref{subsec:datasets}), the preprocessing and dimensionality reduction techniques applied (Section \ref{subsec:dimensionality-reduction}), the classification models used (Section \ref{subsec:classification-models}), the validation techniques (Section \ref{subsec:validation}), and the metrics used to evaluate the performance of the models (Section \ref{subsec:performance-metrics}).

%----------------------------------------------------
\subsection{Dataset} \label{subsec:datasets}
%----------------------------------------------------
The dataset consists of voice recordings at 44.1 kHz of different individuals pronouncing the vowel \textit{/a/} in a prolonged manner since the pronunciation of this vowel present significant differences between healthy individuals and patients with PD \cite{solana2020automatic}. Subsequently, they applied different signal processing algorithms to the recordings to extract useful clinical information. This dataset is made up of 756 samples from 252 individuals, and presents certain particularities:
\begin{itemize}
    \item The data set is three samples for each individual, so the selection of the appropriate validation technique is very important.
    \item Of these 252 individuals, only 64 are healthy individuals, which means that the data set is clearly unbalanced. For this reason, we must be careful with the metrics we use to measure the performance of the models, as biased metrics could lead us to incorrect conclusions.
    \item The dataset has 754 features, but one is the patient identifier and it was removed. Then, the first one is the patient's gender, and the rest belong to the feature types defined in Section \ref{sec:voice-recordings}, the specific number of characteristics of each group is offered in Table \ref{tab:features-dataset}. The ratio samples/features is almost equal to one, which highlights the need to perform a previous dimensionality reduction step so that we have a more manageable set and more explanatory results.
\end{itemize}

\begin{table}[]
\centering
\caption{Groups of features in the dataset.}
\label{tab:features-dataset}
\begin{tabular}{cl|c}
\hline
\multicolumn{1}{|c|}{\textbf{Group}}                                                                                            & \multicolumn{1}{c|}{\textbf{Measure}}                                                      & \multicolumn{1}{c|}{\textbf{\#features}} \\ \hline
\multicolumn{1}{|c|}{\multirow{9}{*}{\textbf{G1}}}                                  & Jitter variants & \multicolumn{1}{c|}{5} \\ \cline{2-3} 
\multicolumn{1}{|c|}{} & Shimmer variants & \multicolumn{1}{c|}{6} \\\cline{2-3} 
\multicolumn{1}{|c|}{} & Fundamental frequency parameters & \multicolumn{1}{c|}{5}  \\ \cline{2-3} 
\multicolumn{1}{|c|}{} & Harmonicity parameters & \multicolumn{1}{c|}{2}  \\\cline{2-3} 
\multicolumn{1}{|c|}{} & Recurrence Period Density Entropy & \multicolumn{1}{c|}{1} \\\cline{2-3} 
\multicolumn{1}{|c|}{} & \begin{tabular}[c]{@{}l@{}} Detrended Fluctuation Analysis \\ \end{tabular} & \multicolumn{1}{c|}{1} \\\cline{2-3} 
\multicolumn{1}{|c|}{} & Pitch Period Entropy & \multicolumn{1}{c|}{1}\\ \hline
\multicolumn{1}{l}{} & \multicolumn{1}{r|}{\textbf{Total}} & \multicolumn{1}{c}{\textbf{21}}  \\ \hline
\multicolumn{1}{|c|}{\multirow{3}{*}{\textbf{G2}}} & Intensity parameters  & \multicolumn{1}{c|}{3} \\ \cline{2-3} 
\multicolumn{1}{|c|}{} & Formant frequencies & \multicolumn{1}{c|}{4} \\ \cline{2-3} 
\multicolumn{1}{|c|}{}  & Bandwidth & \multicolumn{1}{c|}{4}           \\ \hline
\multicolumn{1}{l}{} & \multicolumn{1}{r|}{\textbf{Total}} & \multicolumn{1}{c}{\textbf{11}}  \\ \hline
\multicolumn{1}{|c|}{\centering\begin{tabular}[c]{@{}l@{}}\textbf{G3}\end{tabular}} & MFCCs  & \multicolumn{1}{c|}{84} \\ \hline
\multicolumn{1}{l}{} & \multicolumn{1}{r|}{\textbf{Total}} & \multicolumn{1}{c}{\textbf{84}}  \\ \hline
\multicolumn{1}{|c|}{\begin{tabular}[c]{@{}l@{}}\textbf{G4}\end{tabular}} & WT  & \multicolumn{1}{c|}{182}  \\ \hline
\multicolumn{1}{l}{} & \multicolumn{1}{r|}{\textbf{Total}} & \multicolumn{1}{c}{\textbf{182}} \\ \hline
\multicolumn{1}{|c|}{\multirow{6}{*}{\textbf{G5}}} & Glottis Quotient & \multicolumn{1}{c|}{3} \\ \cline{2-3} 
\multicolumn{1}{|c|}{} & Glottal to Noise Excitation & \multicolumn{1}{c|}{6} \\ \cline{2-3} 
\multicolumn{1}{|c|}{} & \begin{tabular}[c]{@{}l@{}}Vocal Fold Excitation Ratio \end{tabular}  & \multicolumn{1}{c|}{7}           \\ \cline{2-3} 
\multicolumn{1}{|c|}{} & Empirical Mode Decomposition & \multicolumn{1}{c|}{6} \\ \hline
\multicolumn{1}{l}{} & \multicolumn{1}{r|}{\textbf{Total}}  & \multicolumn{1}{c}{\textbf{22}}\\ \hline
\multicolumn{1}{|c|}{\centering\begin{tabular}[c]{@{}l@{}}\textbf{G6}\end{tabular}} & TQWT & \multicolumn{1}{c|}{432}  \\ \hline
\multicolumn{1}{l}{}  & \multicolumn{1}{r}{\textbf{Total}} & \multicolumn{1}{c}{\textbf{432}}
\end{tabular}
\end{table}

%----------------------------------------------------
\subsection{Dimensionality reduction} \label{subsec:dimensionality-reduction}
%----------------------------------------------------
A dimensionality reduction step is carried out, instigated by the studies analyzed in the state of the art and by the characteristics of the dataset. There are two types of approaches to dimensionality reduction. On the one hand, there is FS, which reduces the number of initial features, keeping those that the algorithms determine are most relevant. On the other hand, we have feature extraction algorithms, which create new features by combining existing ones. Specifically, this work will focus on FS to favor the explainability of the results.  A wide number of selection methods were chosen, ranging from filtering methods such as $\chi^{2}$ or the Pearson correlation coefficient (PCC) where the characteristics are ordered based on these tests to more complex ones that are cited below:

\begin{itemize}
    \item \textit{ReliefF} \cite{urbanowicz2018relief}: 
    weights each feature by comparing each sample with its neighbors. If samples belonging to the same class have a very different value in a feature, the score of the feature goes down. If, on the other hand, the value of the feature remains stable, its weight increases. It is widely used for its noise tolerance.
    \item \textit{Minimum Redundancy Maximum Relevance (mRMR)} \cite{vergara2014review}: orders the features based on how well they are able to precede the response variable based on two criteria: redundancy and relevance. Redundancy refers to the correlation of a given feature with the rest, so that if it is high it loses points in the ranking. Relevance refers to the direct correlation with the output variable, so if it is high it earns points. 
    \item \textit{Term Variance Feature Selection (TVFS)} \cite{theodoridis2006pattern}: calculates the variance of each feature and considers that those with the greatest variance are the ones that contain the most information.
    \item \textit{Correlation Feature Selection (CFS)} \cite{hall1999correlation}: evaluates each feature based on a function that calculates its correlation with the output variable, obtaining a final subset of the most correlated variables.
   \item \textit{Neighbourhood Components Analysis (NCA)} \cite{goldberger2004neighbourhood}: based on attributing weights to the features in a way that minimizes an objective function that measures the average classification error on the training data using the nearest neighbor concept. NCA itself is a classification algorithm similar to the \textit{k-Nearest Neighbors} algorithm, so it is considered a wrapper method.
    \item \textit{Sequential Feature Selection} \cite{ruckstiess2011sequential}: generates a subset of features by sequentially adding the most promising one using the performance of a machine learning model (generally the same one to be trained later) as a criterion.
\end{itemize}

%----------------------------------------------------
\subsection{Classification models} \label{subsec:classification-models}
%----------------------------------------------------

SVM is one of the most used classification models in this problem (see Section \ref{sec:state-of-art}) achieving good results, however, after testing with different parameter configurations (both kernel and C), the results obtained were quite mediocre and, therefore, we focused our efforts on ANNs. For these, a feed-forward network with a single hidden layer and a varying number of units was used. Different training algorithms were also applied to compute the network weights. The rest of the parameters were left as defaults. Table \ref{tab:experimentation-ann} shows the parameter variations made.

\begin{table}[]
\caption{Variations of parameters for ANN experimentation.}
\label{tab:experimentation-ann}
\centering
\begin{tabular}{|c|l|}
\hline
\textbf{Parameter} & \multicolumn{1}{c|}{\textbf{Value}}   \\ \hline
                   & Levenberg-Marquardt (LM) \\ \cline{2-2}                           & Resilient propagation (RP) \\ \cline{2-2} 
                   & BFGS quasi-Newton (BFG) \\ \cline{2-2} 
                   & Scale Conjugate Gradient (SCG) \\ \cline{2-2} 
                   & \begin{tabular}[c]{@{}l@{}} Conjugate gradient with \\ Powell-Beale restarts (CGB)\end{tabular} \\ \cline{2-2} 
                   & \begin{tabular}[c]{@{}l@{}}Fletcher-Powell \\ conjugate gradient (CGF)\end{tabular} \\ \cline{2-2} 
                   & \begin{tabular}[c]{@{}l@{}}Polak-Ribière  \\ conjugate gradient (CGP)\end{tabular} \\ \cline{2-2} 
                   & Gradient descent (GD) \\ \cline{2-2} 
                   & \begin{tabular}[c]{@{}l@{}} Gradient descent with adaptive \\ learning rate algorithm (GDX)\end{tabular} \\ \cline{2-2} 
\multirow{-11}{*}{\begin{tabular}[c]{@{}c@{}}Learning \\ algorithm\end{tabular}} & \begin{tabular}[c]{@{}l@{}} Gradient descent \\ with momentum (GDM)\end{tabular} \\ \hline
\multicolumn{1}{|l|}{} & \multicolumn{1}{c|}{10} \\ \cline{2-2} 
\multicolumn{1}{|l|}{$\#$Hidden Units} & \multicolumn{1}{c|}{50}\\ \hline
\end{tabular}
\end{table}

%----------------------------------------------------
\subsection{Experiment flow}
%----------------------------------------------------

\begin{figure}\centering
    \includegraphics[scale=0.32]{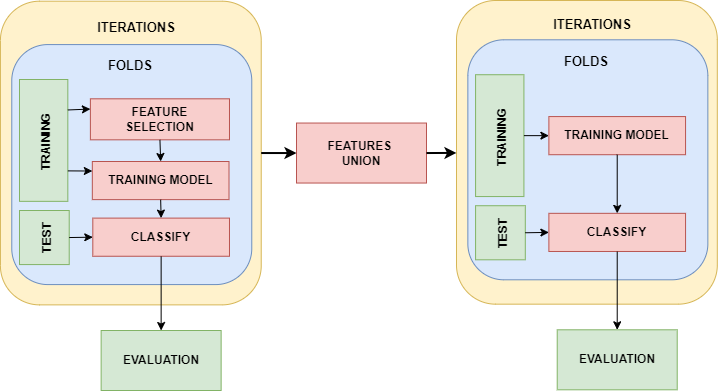}
    \caption{Experiment flow.}
    \label{fig:flow-experiment}
\end{figure}

In this section, we will explain how the stages of FS and subsequent classification have been developed. It is important to note that, the feature set must be unique to facilitate the work of clinical specialists. This has motivated the two-step development illustrated in Fig. \ref{fig:flow-experiment}. Classical machine learning methodologies would follow a similar approach to the left side of the figure, where the training is divided into separate folds and repeated several times. However, this approach would result in many subsets of features (iterations $\times$ folds) and provide, as FS result, the mean of the features used that is not clinically relevant. For this reason, we have chosen to carry out a joining process that integrates the most used features and, with that feature set fixed, it is retrained to obtain the performance results (right part of Fig. \ref{fig:flow-experiment}). 

Performing a union, without any restrictions, can result in very large feature subsets. However, an experimental analysis determined that between 30 and 50 features were enough to achieve good performance results (without sacrificing computational cost).  Therefore, we implement a feature union computation algorithm that uses a threshold (\textit{t}) to decide whether to add a feature to the union set. This threshold only indicates, in percentage terms, the number of times that a feature was chosen in the total number of runs (iterations $\times$ folds). Initially, all features that have been used at least once are considered. If the feature set does not have a size greater than 50, the procedure ends. If it exceeds it, the threshold is increased and those features under the threshold are discarded, and so on until a number of features between 30 and 50 are retained.

Our initial intention was to test this method with all configurations proposed in Section \ref{subsec:classification-models}, as well as for all the FS methods defined in Section \ref{subsec:dimensionality-reduction}. However, ANNs are computationally complex models and require large training times, so we started with a first battery of tests and then continued with variations using those that were considered more promising. For the sake of comparison,
we also trained the different ANNs configurations using the 753 features in the original set (baseline results).

%----------------------------------------------------
\subsection{Validation} \label{subsec:validation}
%----------------------------------------------------
Validation of the models was performed using several approaches. Initially, a 10-fold cross-validation was considered. However, since this dataset consists of three samples per patient (see Section \ref{subsec:datasets}), there is a very high probability that for a given patient, some of the intances are in the test set and others are in the training set \cite{Naranjo2016}, leading to unrealistic performance values. Previous studies used the Leave-One-Person-Out (LOPO) technique, where the three samples belonging to a patient are either in the test set or the training set. However, it does not allow the performance metrics (see Section \ref{subsec:performance-metrics}) to be computed properly without pooling the results. This affects the results obtained, so we have discarded it. Finally, three options were proposed and explicitly implemented:

\begin{enumerate}
    \item \textbf{Grouped K-Fold}: Identical to conventional \textit{K-Fold} cross-validation, except that the three instances belonging to each patient are added together to the test or training set.
    \item \textbf{Stratified Hold-Out by Person}: In each iteration, the dataset is divided into two subsets. The training subset consists of 75\% of the instances belonging to healthy patients (always considering all samples for each patient) and 75\% of the samples belonging to patients with PD. The test subset consists of 25\% of the remaining samples. The samples in each subset are randomly selected. 
    \item \textbf{Summarized LOO} \cite{sakar2013collection}: This technique integrates the three samples from each patient into a single sample. This process can be performed with different metrics (mean, median, etc.). This option was ultimately discarded because it did not seem clinically appropriate and the risk of losing relevant information was too high, in addition to reducing the sample size.
\end{enumerate}

%----------------------------------------------------
\subsection{Performance metrics} \label{subsec:performance-metrics}
%----------------------------------------------------
As we have seen in Section \ref{sec:state-of-art}, many studies use accuracy as one of the performance metrics of the models. However, accuracy is not appropriate in the case of unbalanced ensembles. For this reason, F1-score is used, which is defined as follows

\begin{equation*} F1-score= \frac{2\times precision\times recall}{precision+recall}
\end{equation*}
\begin{equation*}
    precision=\frac{TP}{TP+FP}
\end{equation*}
\begin{equation*}
    recall=\frac{TP}{TP+FN}
\end{equation*}

\noindent being $TP$, $FP$, and $FN$, the true positives, false positives, and false negatives, respectively, of the contingency table. Since the set is unbalanced and this metric can have differences depending on the class considered as positive or negative, we chose to use the \textit{macro-averaged F1-Score}, which performs the mean taking each of the classes as positive. Moreover, the use of Matthew's correlation coefficient (MCC) is proposed \cite{Chicco2020}:

\begin{equation*}
\label{mcc}
    \frac{TP*TN-FP*FN}{\sqrt{(TP+FP)*(TP+FN)*(TN+FP)*(TN+FN)}}
\end{equation*}
which ranges from -1 to 1, with 1 being desirable.
Finally, statistical tests are carried out to select the best learning models.

% ***************************************************
\section{Results and discussion} \label{sec:experimental-results}
% ***************************************************
We started experimenting with SVM because, according to the literature, it was one of the most widely used techniques and the results were expected to be promising. However, after testing 116 combinations of parameters for SVM, the results obtained were rather unsatisfactory. For this reason, this section presents and discusses the results obtained for the ANN models. In addition, the selected features will also be analyzed. After testing the validation techniques mentioned previously, a stratified hold-out by person was chosen in both stages, repeating it 30 times (folds=1, iterations=30).

% ---------------------------------------------------------
\subsection{ANN performance results}
\label{sec::ann_performance_results}
% ---------------------------------------------------------
In the case of the experimentation with ANNs, a total of 60 runs were performed varying the different parameters. Table \ref{fig:comparison-training-algorithms} shows the results obtained for the F1-Score and MCC metrics using the different training algorithms presented in Section \ref{subsec:classification-models}. These results justify the use of the Levenberg-Marquardt (LM) function in the successive phases of experimentation, since, as can be seen, this second-order algorithm obtains significantly better results than the rest.

 \begin{table}[]\centering
 \caption{Comparison of training algorithms for ANN experimentation.}
 \label{fig:comparison-training-algorithms}
 \begin{tabular}{|c|l|l|}
 \hline
 \textbf{Learning Algorithm} & \multicolumn{1}{c|}{\textbf{F1-Score}} & \multicolumn{1}{c|}{\textbf{MCC}} \\ \hline
     BFG & 0.868 +/- 0.057 & 0.746 +/- 0.111 \\ \hline
     CGB & 0.848 +/- 0.054 & 0.707 +/- 0.102 \\ \hline
     CGF & 0.913 +/- 0.045 & 0.832 +/- 0.085 \\ \hline
     CGP & 0.857 +/- 0.063 & 0.726 +/- 0.119 \\ \hline
     GDM & 0.823 +/- 0.083 & 0.667 +/- 0.139 \\ \hline
     GDX & 0.812 +/- 0.051 & 0.645 +/- 0.094 \\ \hline
     GD  & 0.849 +/- 0.050 & 0.712 +/- 0.096 \\ \hline
     LM  & \textbf{0.976 +/- 0.030} & \textbf{0.952 +/- 0.058} \\ \hline
     RP  & 0.714 +/- 0.086 & 0.512 +/- 0.145 \\ \hline
     SCG & 0.828 +/- 0.052 & 0.673 +/- 0.099 \\ \hline
 \end{tabular}
 \end{table}

\begin{table*}[htp]
\caption{Summary of the test results for the ANN. Feat stands for feature.}
\label{tab:ann-results}
\centering
\begin{tabular}{|c|c|c|c|c|c|}
\hline
\textbf{ANN} &\multicolumn{2}{c|}{\textbf{Feat Selection}}&\multicolumn{3}{c|}{\textbf{Performance results}}\\\hline
\textbf{$\#$units hidden layer } & \textbf{Method} & \textbf{$\#$Selected feat} & \textbf{F1-Score} & \textbf{MCC} & \textbf{\#total feat} \\ \hline
10 & baseline & - & 0.968 +/- 0.040 & 0.937 +/- 0.079 &  753 \\ \hline
50 & NCA      & 50  & \textbf{0.987 +/- 0.024} & \textbf{0.974 +/- 0.047} & 48 \\ \hline
75 & NCA & 50 & 0.983 +/- 0.027 & 0.966 +/- 0.053 & 47 \\ \hline
50 & NCA & 20 & 0.979 +/- 0.027 & 0.959 +/- 0.053 & 39 \\ \hline
50 & ReliefF & 20 & 0.979 +/- 0.028 & 0.959 +/- 0.044 & 44 \\ \hline
10 & NCA & 20 & 0.969 +/- 0.030 & 0.938 +/- 0.060 & 38 \\ \hline
10 & NCA & 50  & 0.964 +/- 0.036 & 0.928 +/- 0.072 & 50 \\ \hline
50 & ReliefF & 50 & 0.961 +/- 0.037 & 0.924 +/- 0.072 & 49 \\ \hline
75 & ReliefF & 20  & 0.953 +/- 0.033 & 0.906 +/- 0.064 & 35 \\\hline
50 & $\chi^2$ & 50 & 0.930 +/- 0.035 & 0.863 +/- 0.069 & 42 \\\hline
50 & PCC & 50 & 0.913 +/- 0.042 & 0.828 +/- 0.085 & 49 \\ \hline
75 & mRMR & 50 & 0.911 +/- 0.045 & 0.825 +/- 0.089 & 49 \\ \hline
75 & PCC & 50 & 0.905 +/- 0.046 & 0.819 +/- 0.082 & 46 \\ \hline
50 & mRMR & 50 & 0.907 +/- 0.040 & 0.817 +/- 0.078 & 29 \\ \hline
10 & $\chi^2$ & 50 & 0.895 +/- 0.045 & 0.792 +/- 0.088 & 43 \\\hline
10 & ReliefF & 20 & 0.893 +/- 0.042 & 0.790 +/- 0.083 & 36 \\\hline
\end{tabular}
\end{table*}

The number of combinations run, FS and ANNs with LM algorithm, was very high, to summarize, the 15 best MCC results after the joining process are shown in Table \ref{tab:ann-results}, to facilitate comparisons, results without selection (baseline), are also included. Note that FS methods usually provide a rank of features where a threshold (\emph{top}) must be established to select the top ones, this is the parameter shown in the third column of Table \ref{tab:ann-results}.

We consider that the results obtained for the trained configurations are very satisfactory. In general, the MCC metric presents a rather high value, higher than the baseline in the first five models. The rest of the models, even if they do not exceed the baseline, obtain competent results if we take into account the significantly reduced number of features used in their training. Similarly, the F1-score is also higher in 5 combinations using FS than baseline.

Regarding the configurations used, we see that satisfactory results can be obtained with as few as 10 neurons in the hidden layer, i.e. networks with not very complex architectures. In fact, in general, models with a higher number of hidden units, 75, obtain lower results. 

Among the FS methods studied, we see that NCA obtains the best results. The configuration using an ANN with 50 neurons in the hidden layer and NCA with 50 selected features obtains an MCC value of approximately 0.974, which is better than the baseline (0.937). The second and third best-performing configurations also used NCA, varying the number of neurons and the number of selected features. In addition, NCA is also the most represented FS method among the top 15 models (5 out of 15, 33\%). The ReliefF algorithm is also in second place, with the first of its models outperforming the baseline with an MCC of about 0.959.  ReliefF is the second most represented algorithm (4 out of 15, 27\%). Finally, despite its simplicity,  methods based on $\chi^2$ and PCC, along with mRMR are equally represented (2 out of 15, 13\%).

A multiple comparison test was performed to test for significant differences between the models. NCA, ReliefF, and baseline show no significant differences between them. However, the remaining FS algorithms achieved significantly lower performance.

It is worth mentioning that a fair comparative study with all proposals in Section \ref{sec:state-of-art} is not possible. First, the validation technique used, although adapted for the dataset so that the same patient always goes to training or testing and not both, is not the one used by previous studies. Second, the performance metric does not match either because we opted for using the F1-score because of the unbalanced dataset. Keeping these considerations in mind, when focusing on those studies that use the MCC metric, the performance achieved in this work is much higher than the best of the previous studies (0.974 vs. 0.868).

% ---------------------------------------------------------------------------------
\subsection{Analysis of selected characteristics for ANN} \label{sec:ann-features}
% ---------------------------------------------------------------------------------

\begin{figure*}
    \centering
    \begin{subfigure}[b]{7.0cm}            
        \frame{\includegraphics[scale=0.4]{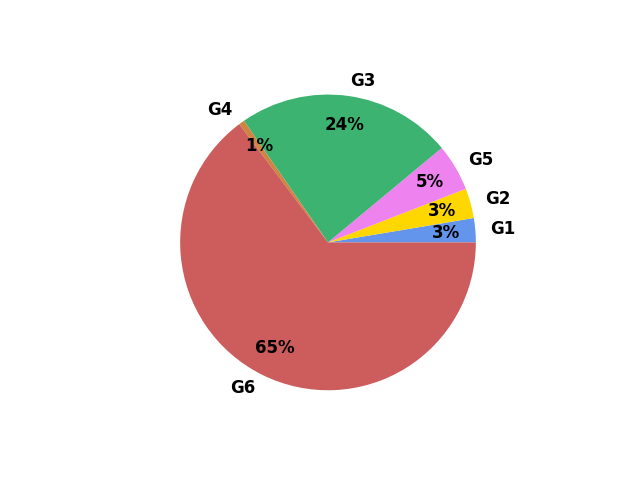}}
        \caption{All features that have ever been selected.}
        \label{fig:selected_features_top15_pie_all_ann}
    \end{subfigure}
    \hspace{1cm}
    \begin{subfigure}[b]{7.0cm}
        \centering
        \frame{\includegraphics[scale=0.4]{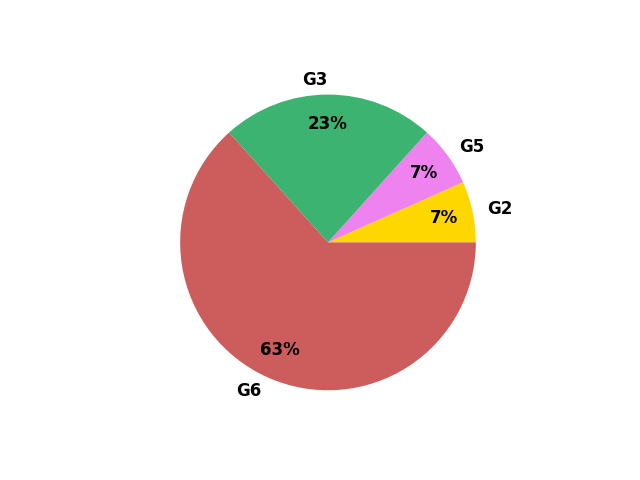}}
        \caption{ The first 30 features.}
        \label{fig:selected_features_top15_pie_top30_ann}
    \end{subfigure}
    \caption{Distribution of selected features in each group in the top 15 ANN models.}
     \label{fig:selected_features_top15_ann}
\end{figure*}

Fig. \ref{fig:selected_features_top15_ann} shows pie charts representing in a percentage way the number of features selected by the top 15 models.  Specifically Fig. \ref{fig:selected_features_top15_pie_all_ann} considers all features that have ever been selected whereas  Fig. \ref{fig:selected_features_top15_pie_top30_ann} take into account only the top 30 selected features. Both graphs show that the feature groups most represented in the top 15 models are G3 (Mel Frequency Cepstral Coefficients, MFCC) and G6 (Tunable Q-factor wavelet transform, TQWT).

\begin{figure*}
    \centering
    \begin{subfigure}[b]{7.0cm}        
        \frame{\includegraphics[scale=0.4]{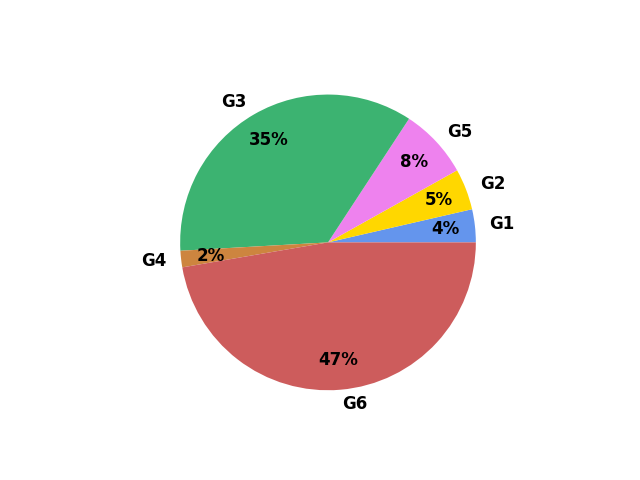}}
        \caption{NCA}
        \label{fig:selected_features_nca_pie_all_ann}
    \end{subfigure}
    \hspace{1cm}
    \begin{subfigure}[b]{7.0cm}
        \centering
        \frame{\includegraphics[scale=0.4]{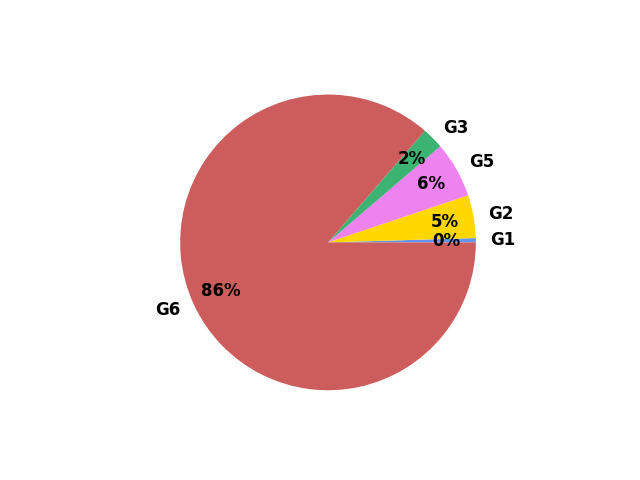}}
        \caption{ReliefF}
        \label{fig:selected_features_relieff_pie_all_ann}
    \end{subfigure}
    
    \caption{Distribution of selected features for the best FS algorithms.}
    \label{fig:selected_features_algorithms_ann}
\end{figure*}

On the other hand, Fig. 3 shows which groups of characteristics have the most representativeness when using the two selection methods that exhibited the best performance, improving on the baseline: NCA (see Fig. \ref{fig:selected_features_nca_pie_all_ann}) and ReliefF (see Fig. \ref{fig:selected_features_relieff_pie_all_ann}). It can be seen that the NCA method selects a greater percentage of features from group G3, while ReliefF leaves them in the background, focusing more on features from group G6.

% ***************************************************
\section{Conclusions} \label{sec:conclusions}
% ***************************************************
The main objective of this work was to classify patients with PD and healthy patients. For this purpose, sets of voice recordings are used as they allow an early and reliable diagnosis. These sets usually have few samples and a high number of features, which makes the learning process difficult and generates classification models with poor generalization capacity. The dataset used consists of 753 features and 756 samples, and dimensionality reduction proved to be very useful. Other particularities of the dataset include three samples per patient and a clear imbalance. Therefore, it was necessary to analyze validation methods and metrics to adequately evaluate the models. In terms of validation metrics, several options were proposed and implemented, of which the stratified holdout by person was the most consistent and realistic.

In terms of results, SVMs do not obtain satisfactory results and show poor performance (MCC value of 0.547), while the ANNs provide a good approach to handling the problem (the best of the models reaches a MCC value of 0.974). For this type of model, the FS algorithms with significantly better classification results were \textit{NCA} and \textit{ReliefF}, both with drastic reductions in the number of features used. The best configuration using NCA only uses 48 features, i.e., 6.4\% of the total features. Similarly, ReliefF employes even less 44 (5.8\%). Therefore, although the statistical tests indicate that there are no differences in performance (MCC) with respect to using the complete set, they clearly exist with respect to the number of features used. A fair comparative study with previous researchs was not possible because of the different metrics and validation methods used. However, some studies do use MCC and compared to them, our results are clearly better (with an increase of 10\%).

Finally, it is worth mentioning which are the features that the best models choose. MFCC (G3) and TQWT (G6) groups are the ones that provided the most information when solving the problem with the studied configurations. In the case of the NCA method, features from both groups were similarly represented, while ReliefF showed a clear preference for features from group G6.

%\textcolor{red}{ESTO TENDRÍAMOS QUE COMENTARLO PARA EXPLICARLO DE FORMA QUE NO NOS DEMOS UN TIRO EN EL PIE Por otro lado, los resultados obtenidos indicaron que el uso de la unión de características seleccionadas a lo largo de las diferentes iteraciones permitió obtener métricas de rendimiento con valores superiores y más robustas. Sin embargo, creemos que estos resultados han de ser estudiados con más profundidad en trabajos posteriores, utilizando diferentes métodos de validación e incluso validando con otros conjuntos de datos con el fin de confirmar esta tendencia.}

As future work, it could be desirable to study the specific features most selected, not only the groups, and their effect on the diagnosis. It could also analyze more complex selection methods, such as wrappers, that use the ANNs themselves as classifiers to see if these results can be improved in any sense, both in performance and feature reduction. Furthermore, in favor of explainability, one can opt for classification methods such as decision trees that would provide an output that is clearly interpretable by clinicians, although perhaps with lower performance.

\section*{Acknowledgments}
This work has been supported by the National Plan for Scientific and Technical Research and Innovation of the Spanish Government (PID2019-109238GB-C22); and by the Xunta de Galicia (ED431C 2022/44) with the European Union ERDF funds. CITIC, as a Research Center of the University System of Galicia, is funded by Conseller\'ia de Educaci\'on,  Universidade e Formaci\'on Profesional of the Xunta de Galicia through the European Regional Development Fund (ERDF) and the Secretar\'ia  Xeral de Universidades (Ref. ED431G 2019/01).

\bibliographystyle{unsrt}  
\bibliography{biblio} 

\end{document}